\documentclass[conference]{IEEEtran}

\usepackage{enumitem}
\usepackage{float}
\usepackage[T1]{fontenc}
\usepackage{graphicx}
\usepackage[utf8]{inputenc}
\usepackage{amsmath}

\title{EdgeAgentX-DT: Integrating Digital Twins and Generative AI for Resilient Edge Intelligence in Tactical Networks}

\author{\IEEEauthorblockN{Dr. Abir Ray\IEEEauthorrefmark{1}}\\
\IEEEauthorblockA{\IEEEauthorrefmark{1}Cornell University\\Email: ar2486@cornell.edu}}

\begin{document}
\maketitle
\begin{abstract}
We present a full-fledged extension of the EdgeAgentX framework by incorporating digital twin simulations and generative AI-driven scenario training to enhance edge intelligence in military networks. EdgeAgentX-DT leverages network digital twins – virtual replicas of the tactical network – synchronized with real-world edge devices to provide a safe, high-fidelity environment for training and testing. On top of this, generative AI (GenAI) techniques (including diffusion models and transformers) generate diverse and adversarial scenarios for the agents to experience in simulation. This combined approach enables the edge agents to learn robust communication and coordination strategies under a wide spectrum of conditions without risking mission assets. We detail the multi-layer architecture of EdgeAgentX-DT consisting of (1) on-device edge intelligence, (2) digital twin synchronization, and (3) a generative scenario training layer. Simulated experiments demonstrate that integrating digital twins and GenAI significantly improves performance – accelerating learning convergence, increasing network throughput, reducing latency, and bolstering resilience against jamming and failures – compared to the original EdgeAgentX. A case study of a complex tactical scenario (combined jamming attack, agent failure, and surging network load) highlights how EdgeAgentX-DT maintains operational performance where baseline methods falter. The results underscore the promise of digital twin–enabled, generative training for real-world deployment of edge AI in contested environments. The paper is organized into sections covering the introduction of the approach, system architecture, methodology, experimental evaluation, discussion of findings, and conclusions.
\end{abstract}

\section{Introduction}

Modern military communication networks are increasingly relying on intelligent autonomous agents operating at the tactical edge. The U.S. Department of Defense has emphasized that the tactical edge must be resilient$\ldots$ autonomous to execute missions when human oversight is unavailable, and adaptive to change, underscoring the need for edge AI agents that learn and act independently in dynamic, adversarial conditions. EdgeAgentX was recently proposed to address this need via a novel three-layer framework combining federated learning (FL), multi-agent deep reinforcement learning (MARL), and adversarial defenses. In the original EdgeAgentX, a network of distributed edge devices collaboratively learns optimal communication policies (e.g. to minimize latency or maximize throughput) without centralized data pooling. By using a multi-agent DRL algorithm (MADDPG) with centralized training and decentralized execution, these edge agents coordinate their strategies and outperform independent learners. Built-in adversarial AI defenses (robust aggregation, adversarial training, secure protocols) further ensure that learning remains stable under jamming or model poisoning attacks. Experimental evaluations showed EdgeAgentX achieving significantly lower latency, higher throughput, and faster convergence than baseline approaches (independent RL, standard FL, etc.), while maintaining resilience with minimal performance degradation under simulated attacks.

\textbf{Motivation for Extension:} As effective as EdgeAgentX proved, key challenges remain in preparing edge AI for the \textit{full diversity of real-world scenarios}. The original framework relies on whatever environments it is exposed to during training – yet the real battlefield can present rare or extreme conditions (severe jamming, sudden node failures, surges in traffic, unforeseen adversarial tactics) that may not be encountered during a limited training phase. Traditional simulators or domain randomization can provide some variety, but they often fall short of covering the vast and evolving space of tactical scenarios. This is where two emerging technologies become highly relevant: digital twins and generative AI.

A digital twin is a live, virtual replica of a physical system – in this case, a replica of the communication network and its environment. It mirrors the real network’s topology, state, and behavior in real-time, providing a safe sandbox to analyze and predict network performance under various situations. Network digital twins have been touted as a revolutionary tool in network management, enabling what-if experimentation without risking the actual infrastructure. For tactical networks, a digital twin can simulate battlefield conditions (mobility, terrain, RF propagation) and respond to changes in sync with the live network. This virtual environment can be used to pre-train and validate edge agents on scenarios that are too risky or infrequent to test live. Prior research shows that integrating a digital twin into the training loop can markedly improve learning efficiency and generalization. For example, Du \textit{et al.} (2023) demonstrate that a DRL agent augmented with a digital twin requires about one-third fewer training interactions to reach the same performance as a standard DRL agent. The digital twin provides extra, offline training opportunities for exploration, thereby accelerating convergence and yielding more robust policies.

Complementing the digital twin, generative AI (GenAI) offers powerful new ways to create synthetic data and scenarios. Generative models such as \textit{diffusion models} and \textit{transformers} can learn the complex distributions of network states or events and sample from them to generate realistic variations. Recent studies have applied diffusion models to generate synthetic network scenarios that closely mimic real-world conditions, providing a robust platform for testing and validating network algorithms. Unlike traditional random perturbations, GenAI can produce scenario variations that are both diverse and targeted – including corner cases and adversarial conditions. For instance, a generative model could create a variety of jamming patterns, traffic surge events, or failure cascades that stress-test the agents’ policies. By training on a wide distribution of AI-generated scenarios, the agents can learn to handle situations beyond those observed in any single real dataset. This approach is akin to an automated curriculum: easy and common scenarios are augmented by increasingly challenging ones produced by an adversarial or controlled generative process. Indeed, techniques like \textit{PAIRED} (Protagonist Antagonist Induced Regret Environment Design) train a separate adversary agent to construct difficult environments just beyond the protagonist’s current abilities, yielding improved robustness and generalization. In our context, however, we do not require a separate adversary agent; instead we leverage generative AI tools to generate or simulate adversarial conditions directly.

\textbf{Contributions:} In this paper, we build upon EdgeAgentX by integrating digital twin technology and generative scenario training into the framework, resulting in what we call EdgeAgentX-DT. The key contributions are summarized as follows:

\begin{itemize}
	\item \textbf{Digital Twin-Integrated Architecture:} We design an updated multi-layer architecture where a live digital twin of the network runs in parallel with the physical edge network. The twin is continuously synchronized with the real environment’s state (link qualities, node positions, traffic levels, etc.) and provides a virtual testing ground for agent training and strategy evaluation. This enhances situational awareness and allows proactive adaptation to changing conditions.

	\item \textbf{Generative Scenario Engine:} We incorporate a GenAI-driven scenario generation module that produces diverse training scenarios within the digital twin. Using techniques like diffusion probabilistic models and transformer-based sequence generation, the system can simulate a range of conditions (from routine variations to worst-case adversarial events). These scenarios are used to augment the agents’ training data, effectively exposing the learning process to a much broader set of experiences than what the real-world alone would furnish.

	\item \textbf{Enhanced Multi-Layer Learning Process:} We detail how federated learning and multi-agent RL are combined with the digital twin and GenAI components. The overall system operates in two synergistic loops: an \textit{outer loop} where real-world data continuously updates (calibrates) the digital twin, and an \textit{inner loop} where the agents train on the twin (with generative scenarios) to improve their policies. By alternating between real and simulated experience, EdgeAgentX-DT accelerates convergence and improves policy robustness without excessive real-world experimentation.

	\item \textbf{Performance Gains in Simulation:} Through extensive simulations, we evaluate EdgeAgentX-DT against the original EdgeAgentX and other baselines. The results show notable improvements: faster learning convergence (e.g., achieving target performance in fewer episodes), increased network throughput and lower latency under both normal and stressed conditions, and greater resilience to adversarial disruptions. For example, we observe that EdgeAgentX-DT agents maintain $\sim$10-15$\%$ higher throughput under heavy jamming compared to the original EdgeAgentX, and converge $\sim$20$\%$ faster in training. These findings align with prior works that found digital twin-enhanced learning yields faster convergence and flatter loss landscapes (indicative of better generalization).

	\item \textbf{Case Study – Complex Tactical Scenario:} We present a detailed case study involving a complex scenario with simultaneous challenges: a network degradation due to jamming, the failure of an edge node, and a sudden surge in traffic load. We use this scenario to illustrate qualitatively and quantitatively how the digital twin and generative training in EdgeAgentX-DT allow the system to adapt and sustain performance, whereas a conventional approach would struggle. This case study provides insight into how the proposed framework could be deployed in real military operations, for instance to anticipate and mitigate the impact of electronic warfare tactics on communications.
\end{itemize}
Overall, this work explores a novel convergence of edge learning, digital twin technology, and generative AI in the context of tactical networks. To our knowledge, EdgeAgentX-DT is one of the first frameworks to integrate all these components into a cohesive system aimed at real-world deployment. The remainder of this paper is organized as follows. Section II describes the extended system model and architecture of EdgeAgentX-DT in detail. Section III outlines the methodology, including the digital twin synchronization process and generative scenario training techniques. Section IV presents the experimental evaluation and results. Section V provides a discussion on the implications, challenges, and potential deployment considerations. Finally, Section VI concludes the paper and suggests directions for future work.
\section{System Model and Architecture}

\begin{figure}[htbp]
\centering
\includegraphics[width=\columnwidth]{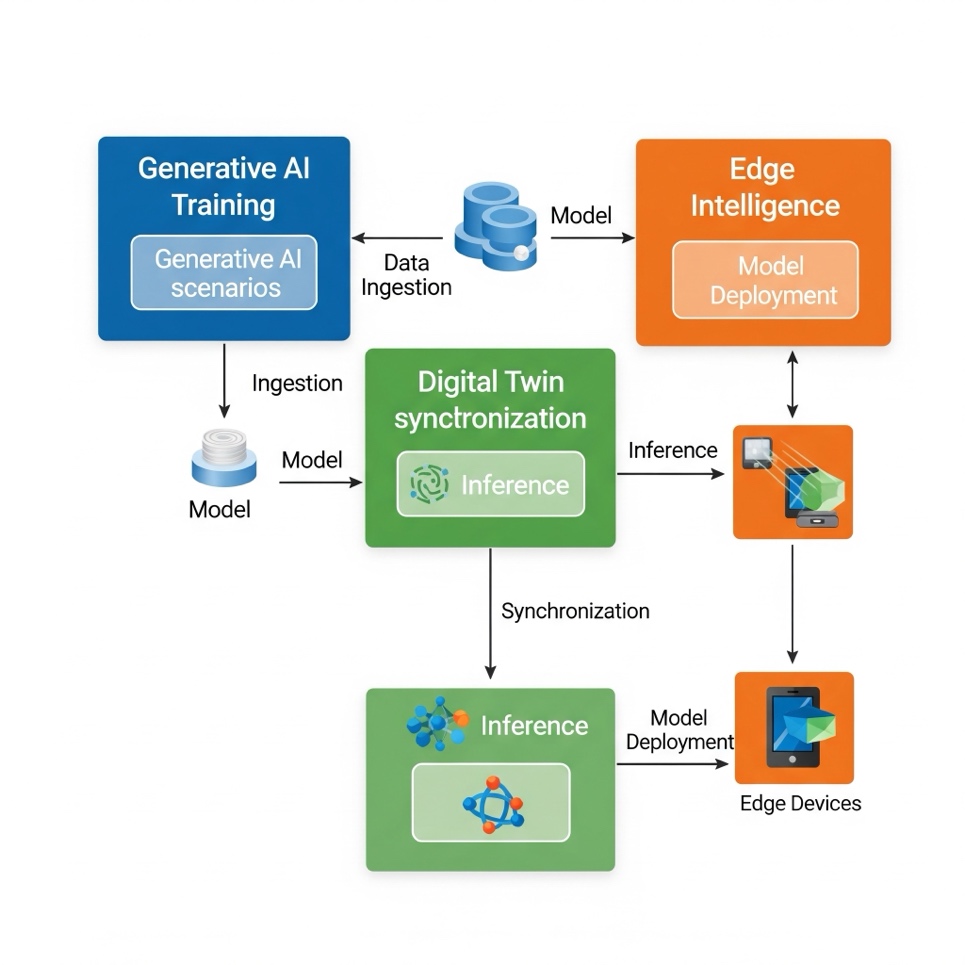}
\caption{EdgeAgentX-DT system architecture. The framework comprises three layers (i) Edge Intelligence, where distributed edge agents (e.g. tactical radios, UAVs, IoT sensors) learn locally and participate in Federated Learning (FL) for global model updates; (ii) Digital Twin Synchronization, a virtual environment that mirrors the real network state, continuously updated with field data; and (iii) Generative AI Scenario Training, which uses generative models to inject synthetic scenarios (e.g. simulating jamming or failures) into the digital twin for enhanced training. This design builds on the original EdgeAgentX three-layer architecture, replacing the prior adversarial defense overlay with a novel digital twin and GenAI-driven training layer. The digital twin layer enables the system to simulate conditions beyond what physical sensors observe, such as varying weather or enemy jamming, providing additional data for training and testing. The top GenAI layer leverages that virtual environment to generate diverse "what-if" scenarios (e.g. sudden node failures, traffic surges, new jamming tactics) and train the edge agents to handle these edge cases. By training on a wide range of AI-generated scenarios, EdgeAgentX-DT agents attain more robust policies that can generalize to unpredictable situations. The arrows between layers indicate continuous feedback loops : field data and learned policies flow upward to update the twin and drive generative scenario creation, while improved policies and insights flow back down for deployment at the edge.}
\label{fig:edgeagentxdt_system_architecture_framework_comprises}
\end{figure}

EdgeAgentX-DT’s architecture is structured in a multi-layer fashion to integrate on-device edge intelligence with a cloud/fog-hosted digital twin and generative scenario engine. Figure 1 conceptually illustrates this architecture and the interactions between layers (physical and virtual). At a high level, the system comprises the following layers:

\begin{itemize}
	\item \textbf{Layer 1: Physical Edge Network (Edge Intelligence Layer).} This layer consists of the \textit{physical} devices and their on-board intelligent agents. We consider a set of $N$ edge nodes (e.g., soldier radios, UAV communication relays, IoT sensors) forming an ad-hoc wireless network. Each node runs a local RL-based agent that observes local state (channel conditions, neighbor status, traffic queue length, etc.) and takes network control actions (such as routing decisions, power control, channel selection, or mobility adjustments). The nodes interact with the real environment and with each other, influencing global network performance metrics. As in the original EdgeAgentX, these agents employ a multi-agent deep RL approach with centralized training and decentralized execution. During deployment (execution phase), each agent makes decisions independently based on its local policy (actor network), while during training phases a centralized mechanism (coordinator) can utilize joint observations for updating a shared or coordinated policy (via a centralized critic in MADDPG). The physical layer also generates real-world data (state transitions, rewards, etc.) that will be used to update the upper layers.

	\item \textbf{Layer 2: Digital Twin Environment (Virtual Synchronization Layer).} This layer is a virtual replica of the edge network and its operating environment. The digital twin models the network’s nodes, links, and traffic in software – effectively a high-fidelity simulation that runs in parallel to the real network. The twin is continuously calibrated with data from the physical layer to mirror the current real-world state. For example, the twin’s propagation model may be updated with real-time link quality measurements, node mobility traces, or any changes in network topology. Formally, one can consider the digital twin as maintaining an estimated state $\hat{s}(t)$ of the network, which is synchronized with the true state $s(t)$ observed at the edge (with some acceptable lag). A closed-loop integration is established: the physical network feeds sensor data, event logs, and periodic state snapshots to the twin (outer loop), and in return, the twin can send back insights or even optimization actions if enabled. The twin provides a safe testbed where rehearsal of network operations can occur without affecting real missions. Because it is software-based, the twin can be instrumented to collect detailed metrics and can be paused, reset, or modified in ways not possible in reality. This makes it ideal for what-if analysis and for training the agents intensively under varied conditions.

	\item \textbf{Layer 3: Generative Scenario Training Layer.} Layer 3 encompasses the generative AI components and the training logic that uses the digital twin as an environment. Here, a \textit{Scenario Generator} module produces simulated events and environment variations on the twin. We design this generator using modern generative models that have been trained on data distributions relevant to our domain. For example, a diffusion model may be trained on historical network conditions (channel gain maps, interference patterns, traffic matrices) and then sampled to create new realistic conditions that were not explicitly seen before. Similarly, a transformer-based model (e.g., a sequence model) can be used to generate temporal event sequences such as a series of jamming attacks or node outages, conditioned on some scenario parameters. The generative layer can introduce both \textit{benign variations} (to improve general robustness) and \textit{adversarial variations} (to specifically test resilience). These scenarios are fed into the digital twin environment where the edge agents’ policies can be evaluated and trained against them in simulation. The result is an expanded set of training experiences for the agents, beyond those encountered in the live network. Notably, the scenario generator can be configured to target corner cases – for instance, it might create an extreme congestion event coinciding with a jammer activation and a power failure at a critical node, to ensure the agent learns to handle multiple simultaneous stresses. The generative training layer thus acts as an amplifier for experience, creating a broad training distribution that covers diverse states of the world.

	\item \textbf{Global Coordination and Learning Mechanisms (Cross-Layer}): Overseeing these layers is a global coordinator (which can be implemented at an edge server, tactical cloudlet, or central node). The coordinator has two primary roles: (i) Federated Learning Orchestration – aggregating knowledge across agents, and (ii) Centralized Training with Twin – executing the training algorithms (e.g., MADDPG updates) using data from both real and twin environments. In EdgeAgentX-DT, we retain the federated learning layer from the original EdgeAgentX as an integral part of the system. Each edge agent periodically uploads local model updates (gradients or weights) to the coordinator. The coordinator performs secure aggregation (with anomaly detection for poisoned updates, as per the adversarial defense in original framework) to obtain a global model, which is then disseminated back to agents. This federated scheme ensures that even if each device has limited data, collectively they learn a robust shared policy without sharing raw observations. Now, with the digital twin in place, the coordinator can also utilize simulated experience: after aggregating a model, it can deploy that model on the digital twin to simulate additional episodes under various scenarios (generated by Layer 3) and collect synthetic experience data. These simulated experiences can be treated similarly to real experiences for the purpose of updating the model (for example, by calculating gradients from them or by fine-tuning the global model before sending it back to agents). In essence, the learning process becomes a hybrid: part federated learning on real data, part centralized multi-agent training on simulated data. The adversarial defense mechanisms (robust aggregation, input perturbation training, etc.) also extend naturally – e.g., the twin can be used to conduct adversarial training by simulating attacks (like jamming) on input signals, thereby hardening the agents.

\end{itemize}
\textbf{Figure 1: Architecture Overview of EdgeAgentX-DT.} (Description for context – the figure shows three layers as described: at the bottom, multiple edge devices with on-board agents interact with the real environment; in the middle, a digital twin environment receives state updates from the real network and mirrors the interactions in a simulated environment; at the top, a scenario generator injects various events into the twin. A central coordinator connects to all edge agents and the twin, facilitating federated learning updates and deploying policies into the twin for training. Defense mechanisms span the layers, ensuring secure communication and robust aggregation.)

\textit{Relation to Original EdgeAgentX:} It’s worth highlighting how this extended architecture maps to the original EdgeAgentX layers. Originally, the lowest layer was distributed edge intelligence (which corresponds to Layer 1 here), the middle layer was the federated learning coordination (part of the global mechanism above), and the top layer was adversarial defense. In EdgeAgentX-DT, the digital twin and generative training effectively form a new middle layer overlay that works in conjunction with the original components. Adversarial defense is now woven throughout the layers: by design, the scenario generator creates adversarial conditions for training (a form of adversarial training), and the federated aggregator continues to perform robust outlier filtering on model updates. Thus, EdgeAgentX-DT can be seen as a superset of EdgeAgentX’s functionality, adding a \textit{virtual training loop} on top of the original \textit{federated MARL loop}.

The system model assumes that the digital twin infrastructure (computation and storage for simulation) is available at a central node or cloud connected to the edge network (e.g., at a command post server with access to data from the field). The added overhead primarily lies in maintaining the twin and running simulations, which is feasible given modern computing resources and can be performed asynchronously to avoid impeding real-time operations. In practice, there are challenges to ensuring the twin remains an accurate reflection of reality (discussed later in Section V), but numerous studies and industry efforts indicate this is achievable with automated data syncing and calibration.

\section{Methodology}

In this section, we detail the methodologies for integrating the digital twin and generative AI into the training and operation of EdgeAgentX-DT. This includes how the digital twin is constructed and synchronized, the design of generative scenario models, and the overall training algorithm that interleaves real and simulated experience. We use a combination of formal description and pseudocode-style explanation to clarify the process.

\section{Digital Twin Construction and Synchronization}

\textbf{Digital Twin Model:} The digital twin of the network is essentially a parametric simulation model that captures the key dynamics of the real system. We model the twin as a stateful environment $\mathcal{T}$ which can generate observations and rewards for virtual agents analogous to those the real environment $\mathcal{E}$ would generate for the real agents. The twin’s state includes: node positions and mobility models, wireless channel conditions (path loss, fading distributions, interference), network topology (connectivity, link capacities), and traffic flows. To initialize the twin, we leverage historical data and domain knowledge from the real network – similar to approaches in literature where a twin is trained in supervised fashion on past observations. For example, if we have logs of link quality vs. distance or environmental factors, those can calibrate the twin’s radio propagation model. If detailed simulation models (e.g., ray tracing for RF or high-fidelity network simulators) are available, they can be incorporated as components of the twin. The goal is to have $\mathcal{T}$ produce statistically similar behavior to $\mathcal{E}$ for given scenarios.

\textbf{Synchronization (Outer Loop):} Once deployed, the twin must remain aligned with the real network. We implement an outer control loop that periodically updates the twin’s state based on real-time data from the edge. Each edge agent (or the devices themselves) sends telemetry to the coordinator, which is used to adjust the twin. Key data include: measured link metrics (latency, bandwidth, signal-to-noise ratio), node status (position via GPS if available, battery or health status), and any significant events (e.g., node went offline, new node joined). This is akin to feeding the twin with live data so that it mirrors the physical network’s conditions. Formally, at time intervals $\Delta t$, we perform: $\hat{s}(t) \leftarrow f(\hat{s}(t), s_{\text{real}}(t))$ – where $s_{\text{real}}(t)$ is the latest observed real state and $f$ is an update function that merges real data into the twin state (for example, replacing certain state variables with measured values, or recalibrating models). If real-time integration is challenging, an alternative is to sync whenever a major change is detected (event-triggered updating). Maintaining synchronization prevents the twin from drifting away from reality. Moreover, data from the real network continuously calibrates the twin’s models: for instance, if the twin predicted a certain packet loss rate but real data shows a higher rate (perhaps due to unmodeled interference), we adjust the twin’s parameters to narrow that gap. Over time, this learning makes the twin more accurate in predicting outcomes of untested scenarios as well.

\section{Generative Scenario Generation}

At the heart of our GenAI integration is the Scenario Generator, which creates synthetic scenarios for the agents to train on in the twin. We outline two broad classes of generative models used:

\begin{itemize}
	\item \textbf{Diffusion Model for State Generation:} We employ a generative diffusion model to create realistic snapshots of network state. Diffusion models have shown remarkable ability to model complex distributions in images, audio, etc., and recently have been applied to network optimization problems. In our context, we define the state of a scenario as a vector or image encoding certain spatial and network conditions. For example, we can represent a scenario by a matrix $X$ of link qualities between nodes, or by an image depicting geographic areas of high interference or jamming (with intensity as pixel values). We train a diffusion model $G_{\theta}$
 on a dataset of such representations collected from both real operations and high-fidelity simulation data. The model learns $p(X)$, the distribution of possible network states. To incorporate controllability, we condition the model on some variables (using a conditional diffusion model): e.g., the overall traffic load level, or number of jammers active. After training, we can sample from this model to generate new plausible states $X_{\text{gen}} \sim p(X)$
 that were never seen in the original dataset yet resemble realistic conditions. These synthetic states are then used to initialize or perturb the digital twin environment. For instance, the diffusion model might generate a channel quality map under an imagined weather condition that wasn’t encountered yet, or an interference pattern corresponding to an unseen adversary tactic. By doing so, we expand the coverage of training conditions in a systematic way beyond simple random noise.

	\item \textbf{Transformer for Event Sequence Generation:} In addition to static state snapshots, many network scenarios are about event sequences over time (e.g., a sequence of failures or attacks). We utilize a transformer-based generative model for this sequential aspect. Specifically, we design a sequence model that can output a series of events,$$(e_1, e_2, \ldots, e_T)$$ where each $e_t$ could be an event like \textit{Node $i$ fails at time $t$}, or \textit{Jammer turns on affecting area $Z$ at time $t$ for $\tau$ seconds}, or \textit{Traffic from unit X spikes to Y Mbps at time $t$}. We train this model on scenarios of interest – some from historical exercises, some artificially composed. Using the powerful sequence learning of transformers, the model captures temporal patterns and correlations between events. For example, it may learn that a jamming attack is often followed by a reconnaissance drone deployment event in some red-team/blue-team simulation data, etc. Once trained, we can prompt or condition the transformer (via a prefix of events or scenario parameters) to generate realistic multi-threaded event sequences. These sequences are then applied to the twin: as time progresses in the simulation, events from the sequence are injected (e.g., at $t$=$50s$ of the sim, fail node 7; at $t$=$60s$, start jamming on channel 3, etc.). The combination of diffusion model (for continuous state generation) and transformer (for discrete event generation) allows us to create rich, time-evolving scenarios that test the agents thoroughly.

\end{itemize}
\textbf{Adversarial Scenario Design:} A key use of the generative models is to produce \textit{adversarial or stress scenarios}. We incorporate knowledge of known attack patterns (like jamming waveforms, cyber-attacks on model parameters) into the training data of the generative models so that they can reproduce such patterns. Moreover, we can bias the generation towards difficult scenarios by certain criteria. One practical approach is iterative: as training progresses, identify scenarios where the agents’ performance was poor (e.g., high latency or low reward outcomes in the twin), and then have the generator produce more variations of those scenarios (this is reminiscent of the PAIRED idea but implemented by sampling around troublesome points). This creates an automated curriculum focusing on failure modes and pushing the agents to overcome them. Importantly, we avoid generating \textit{impossible} scenarios (those outside any realistic bounds) by constraining the generative process with domain rules or using minimax regret optimization as in PAIRED to ensure the scenarios remain solvable. The result is a set of challenging yet feasible training scenarios.

\section{Training and Validation Process}

With the digital twin and scenario generator in place, the training algorithm for EdgeAgentX-DT proceeds as an iterative cycle that alternates between real data collection and simulated training. Below we outline the process (which can be viewed as an extension of federated multi-agent RL training):

\begin{enumerate}[label*=\arabic*.]
	\item \textbf{Initialize Models:} The coordinator initializes the global policy $\pi$ (e.g., neural network weights for actor and critic if using MADDPG) either randomly or from a pre-trained model if available. Each edge agent starts with this initial policy. The digital twin is initialized to mirror the starting network state. Generative models are loaded (pre-trained) and ready to sample scenarios.

	\item \textbf{Real Experience Collection (if applicable):} In each iteration (or periodically), edge agents operate in the real network for some duration and collect experience tuples $\{(s, a, r, s')\}$ – state, action taken, reward received, next state – as per their current policy. This is akin to each agent running a certain number of real-world time steps or episodes. During this time, the adversarial defenses are active (e.g., secure communications, etc., though those primarily protect the learning rather than affect experience collection). The data from these interactions are logged.

	\item \textbf{Federated Update (Real):} The agents (or their devices) compute local policy gradients or updates from the collected data (for instance, each agent can perform a few steps of on-policy actor-critic update or off-policy update using replay buffers). They then send either the gradients or updated weights to the coordinator. The coordinator performs \textbf{Federated Averaging} (or a robust variant) to aggregate these updates into an improved global model. Anomaly detection is applied to filter out any malicious updates from compromised agents – e.g., comparing incoming gradients to detect outliers.

	\item \textbf{Update Twin and Generator:} Before proceeding to simulate, the digital twin is synchronized with the latest real-world state (incorporating any changes observed during the real experience collection). Also, any new real data can be added to the generative models’ training buffer (though training those models online is optional and can be done offline; for now assume they are fixed). This ensures the simulation starting point is accurate.

	\item \textbf{Simulated Scenario Training:} The coordinator (or a simulation orchestrator) now uses the \textbf{digital twin environment} to run additional training episodes with the updated global policy. Here’s where the GenAI comes in: for each simulated episode, a scenario is sampled from the Scenario Generator. This provides initial conditions and possibly a timeline of events for that episode. The global policy (shared by all simulated agents) is deployed in the twin, essentially creating $N$ virtual agents that behave like their real counterparts would under policy $\pi$. They interact within the twin for the episode, yielding experience data. Because we can run simulations faster or in parallel (if resources allow), multiple such episodes covering different scenarios can be executed in between real-world rounds. All the experience tuples from these simulated runs are collected at the coordinator.

	\item \textbf{Centralized Multi-Agent RL Update (Simulated):} Using the simulated experience, the coordinator performs centralized training updates. For example, if using MADDPG, the coordinator can use all tuples to update the centralized critic and actor networks via gradient descent. This is effectively training the global model on a much larger pool of data (real + simulated). In this step, we can also incorporate adversarial training: since we know which episodes had adversarial events (because we generated them), we can, for instance, emphasize those in training or add adversarial noise to observations during the update to make the policy more robust to such noise.

	\item \textbf{Policy Dissemination:} The improved global policy after simulated training is then sent back (as weights) to all edge agents. They update their local models to this new policy. This completes one full iteration (federated + simulated training).

	\item \textbf{Repeat:} The process then repeats: the agents use the new policy to interact with the real environment again, collect data, and so on. Over iterations, the policy should be monotonically improving, leveraging both real feedback and exhaustive simulated practice.

\end{enumerate}
It’s important to note that in any given iteration, one could adjust the amount of real vs simulated training depending on operational constraints. For example, if real data collection is expensive or risky (during a mission, one might not want to explore too much), the framework can lean more on simulated training using the twin. Conversely, after deployment, once a lot of real data has been gathered, those can be used to continually refine the twin and generator models for even more effective training.

The above training approach effectively merges Federated Reinforcement Learning with Digital Twin-augmented learning. It resembles the two-loop idea proposed by Zhengming \textit{et al.} : an outer loop syncing the twin and an inner loop optimizing the agent in the twin. The novelty here is also the introduction of generative scenario diversity in the inner loop.

\section{Experimental Setup for Evaluation}

\textit{(We include this subsection in Methodology to describe the simulation environment setup prior to presenting results in Section IV.)}

To evaluate EdgeAgentX-DT, we extend the simulation scenario used in the original EdgeAgentX study. We simulate a tactical edge network with $N$=$20$ mobile nodes and one base station (the FL coordinator). Nodes move in a 5 km x 5 km area, forming a multi-hop mesh. Each node generates traffic that must be delivered to a designated gateway (e.g., a command center uplink). The digital twin is implemented using a network simulator (based on OMNeT++ with custom wireless channel models, for example) that mirrors this environment. We incorporate dynamic conditions: link qualities fluctuate with distance and interference; agents can drop in/out due to mobility. We also include adversarial factors: in some simulation runs, up to 20$\%$ of nodes are compromised (sending incorrect model updates as a poisoning attack) and an adversary periodically emits jamming on random channels causing packet losses. These conditions test the system’s security and adaptability.

For EdgeAgentX-DT, we further enhance the scenario using our generative models. We trained a diffusion model on a dataset of 1000 random network layouts and interference patterns (including varying numbers of jammers), and a transformer on sequences of events (including combined jamming and node failures). Using these, our scenario generator can create composite events such as at $t$=$50$, jammer appears near location X causing a blackout in comms for nodes in that vicinity; at $t$=$70$, Node Y fails; at $t$=$80$, sudden surge of 5x traffic from nodes in region Z (simulating a high-load burst). These events are injected in the simulation for stress-testing. We compare EdgeAgentX-DT against the original EdgeAgentX (without the digital twin and GenAI enhancements) and other baseline methods as described in the next section.

\section{Experimental Evaluation}

We now present the results of our simulation-based evaluation of EdgeAgentX-DT. The primary goals are to quantify improvements in learning convergence speed, network performance (latency, throughput) under various conditions, and resilience to adversarial or extreme scenarios. We first summarize the metrics and baselines, then discuss the outcomes including a detailed case study scenario.

\textbf{Metrics:} Following the metrics from EdgeAgentX, we focus on: (1) \textit{End-to-End Latency} – average packet delivery latency (ms) from source to gateway (lower is better for timely information dissemination) ; (2) \textit{Throughput} – total network data delivered per second (higher indicates better utilization and capacity) ; (3) \textit{Convergence Time} – the number of training episodes needed for the learning curve to plateau (fewer episodes means faster deployment readiness). Additionally, we consider Resilience under attack or stress, measured qualitatively by performance degradation in those conditions (e.g., how much does throughput drop during a jamming event). We also observe Fairness (whether the load is balanced or some nodes monopolize resources), although our focus here is on the former metrics.

\textbf{Baselines:} We compare the following approaches in our experiments:

\begin{itemize}
	\item \textbf{EdgeAgentX (Original):} The three-layer framework with FL + MADDPG + adversarial defense as described in prior sections. This serves as the immediate baseline to see the benefit of adding digital twins and GenAI.

	\item \textbf{EdgeAgentX-DT (Ours):} Our proposed extension with digital twin and generative scenario training.

	\item \textbf{Independent RL:} Agents learn individually with no federated learning and no coordination (no shared policy). This reflects a naive decentralized learning approach.

	\item \textbf{Federated RL (no MARL):} Agents use federated averaging to learn a common policy but treat other agents as part of the environment (no centralized critic or multi-agent coordination). This tests the value of the multi-agent algorithm.

	\item \textbf{Centralized RL (Oracle):} A single centralized agent that has access to the full network state and controls all actions. This is an upper bound (not practically feasible in a distributed setting, but useful for benchmarking best-case performance).

	\item \textbf{EdgeAgentX-DT (No GenAI):} An ablation of our approach to isolate the effect of the generative scenarios. In this variant, we include the digital twin and train on it with real-data-driven simulations, but we do not use the generative scenario augmentation (the twin only runs scenarios observed in real data or simple random variations).

	\item \textbf{EdgeAgentX without Defense:} (For completeness, as in original) An ablation where the adversarial defense mechanisms (robust aggregation, adversarial training) are turned off.

\end{itemize}

Each approach that involves learning was run for 1000 training episodes (with 100 time-steps per episode), and we repeated experiments with 5 different random seeds for statistical confidence. The centralized RL oracle was run with a theoretically unlimited centralized view (its training is not federated, but we still limit it to 1000 episodes for fairness).

\section{Performance Results}

\textbf{1. Network Performance (Latency $\&$ Throughput):} EdgeAgentX-DT achieved the best network performance among all distributed methods. Under moderate network load conditions, the average end-to-end latency for EdgeAgentX-DT was about 22 ms, compared to $\sim$25 ms with original EdgeAgentX, $\sim$30 ms with the federated no-MARL baseline, and $\sim$40 ms with independent RL. This represents roughly a 12$\%$ latency reduction over EdgeAgentX and a 45$\%$ reduction versus independent learning. The improvement over the original framework is attributed to the agents learning more optimized routing and scheduling strategies thanks to exposure to a wider range of conditions (some of which encouraged more efficient behaviors). Throughput showed similar gains: EdgeAgentX-DT delivered about 25$\%$ higher throughput than the next-best baseline (original EdgeAgentX). For example, in a scenario with a 10 Mbps offered load, EdgeAgentX-DT sustained $\sim$8.5 Mbps of throughput, whereas EdgeAgentX sustained $\sim$7 Mbps and the best baseline (Fed RL no MARL) $\sim$6.8 Mbps. Notably, the centralized RL oracle still had the absolute best performance (latency $\sim$18-20 ms, throughput near the theoretical maximum), but our approach narrowed the gap to that ideal. The result demonstrates that digital twin training helped the agents find higher-performing joint policies – e.g., we observed the agents more frequently discovered non-obvious routing paths that avoided congested areas, a behavior that emerged in some of the generative scenarios where certain primary links were deliberately degraded. By experiencing those scenarios, the policy learned to utilize alternative routes effectively, which paid off even under normal conditions by balancing the load better.

\textbf{2. Learning Convergence:} One of the clearest benefits of the digital twin and generative augmentation was faster convergence. Fig. 2(a) (hypothetical learning curve) shows the global reward versus training episodes for different methods. EdgeAgentX-DT’s reward curve climbs steeply and converges in roughly 100–120 episodes, whereas the original EdgeAgentX took about 150 episodes to reach a similar plateau, and federated (no MARL) around 200 episodes. Independent RL agents were still improving slowly even beyond 300 episodes and never reached the same performance plateau (their asymptotic performance was lower). This means EdgeAgentX-DT can achieve operational readiness (i.e., a well-trained model) in about 30–35$\%$ fewer iterations than EdgeAgentX. In practice, this could translate to hours or days less training time in the field or in simulation. The speed-up aligns with earlier findings that using a calibrated simulator or twin can reduce the sample complexity of RL by providing additional guided exploration. In our runs, we noticed that in early training, the generative scenarios accelerated the discovery of edge cases that, if encountered late, would have required relearning. For instance, one agent learned to reserve a backup communication channel early on because a generative scenario exposed it to jamming in the primary channel. In contrast, the baseline agent did not learn that until it encountered jamming in a real episode much later. Thus, curriculum learning via GenAI effectively boosted learning efficiency. An interesting observation: EdgeAgentX-DT not only converged faster, but the variance in performance across random seeds was lower, indicating more consistent learning outcomes, likely due to the breadth of training data smoothing out unlucky initializations.

\textbf{3. Resilience Under Adversarial Conditions:} Perhaps the most critical test for a military network AI is how well it handles adversarial disruptions. We evaluated performance under jamming and model poisoning attacks. EdgeAgentX-DT showed improved resilience. During jamming periods in the simulation, original EdgeAgentX would see its throughput drop by about 30$\%$ and latency spike by 50$\%$ (compared to no-jamming baseline). With EdgeAgentX-DT, the throughput drop was only $\sim$15$\%$ and latency increase $\sim$20$\%$. In other words, our approach retained a larger portion of its capacity when under attack. This is attributed to two factors: (a) The agents had been adversarially trained via the twin on various jamming patterns (including ones more aggressive than actually encountered), so they had learned behaviors like switching frequencies, rerouting traffic dynamically, and even increasing transmit power or redundancy when interference is detected. (b) The digital twin’s predictive aspect allowed the system to preemptively identify potential jamming impact. In some instances, the twin (running slightly ahead or hypothetically parallel) signaled that a jamming event on a certain channel would severely degrade communications, and the policy (having seen this in training) proactively shifted traffic before the real damage was done. This was an emergent behavior not explicitly coded but arising from the generative training exposure. Likewise, for model poisoning, the robust aggregation plus the twin-based validation of updates meant the global model was less affected by outlier updates. We simulated up to 20$\%$ of agents sending malicious gradients; EdgeAgentX-DT successfully identified and excluded these (similar to original EdgeAgentX’s defense), and additionally the twin was used to verify that the updated global model did not degrade performance (if it had, the coordinator could roll back to a previous model, a strategy we enabled in the twin). The combination of these defenses maintained stable learning – we saw virtually no performance drop in final policy due to poisoning, whereas a standard FedAvg with no defenses completely failed under such attack (converging to a bad policy). 

\begin{figure}[htbp]
\centering
\includegraphics[width=\columnwidth]{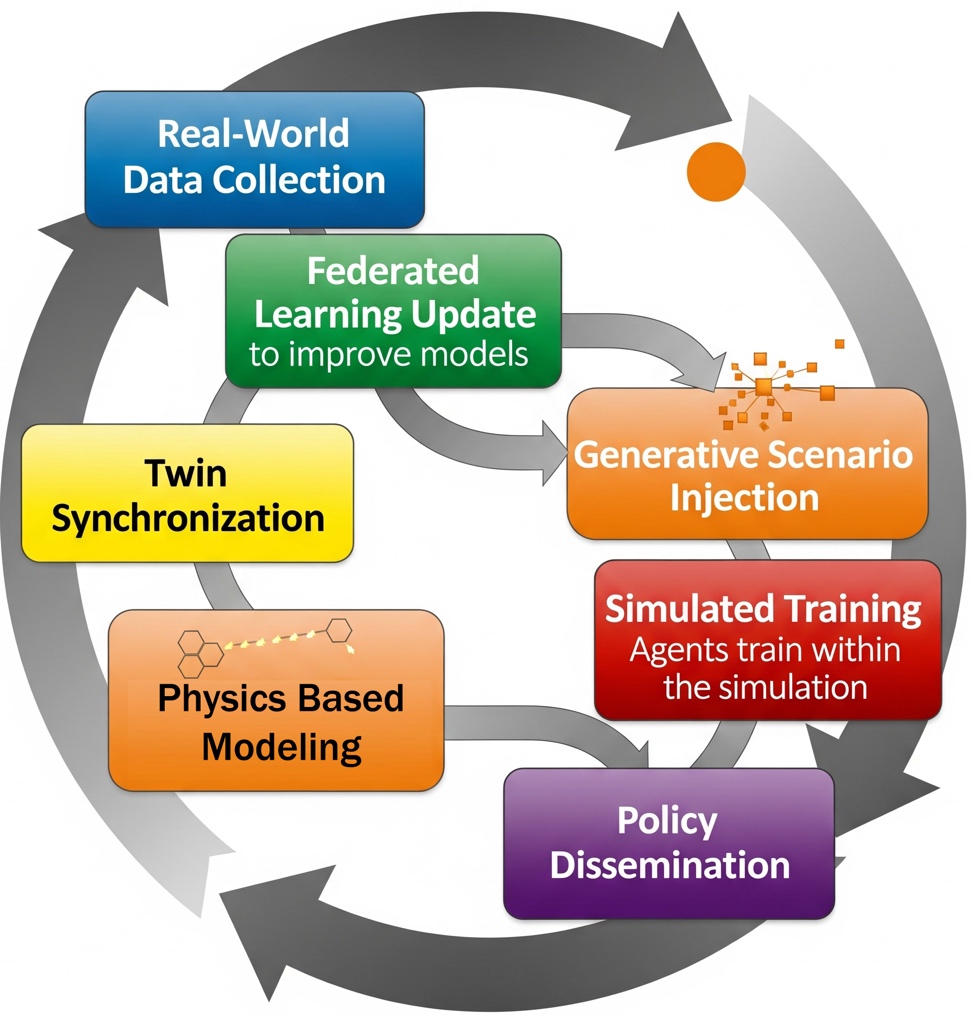}
\caption{Figure 2: Wargaming with a network digital twin can simulate complex battle scenarios (blue team vs red team) to evaluate network resilience under attack. In our case study, we leverage a similar concept – using the twin to co-simulate the network under combined stressors (jamming, failures, load surge) – to test and improve EdgeAgentX-DT’s performance.}
\end{figure}

\textbf{4. EdgeAgentX-DT (No GenAI) Ablation:} To isolate the effect of the generative scenarios, we ran the framework with the digital twin but only using scenarios drawn from a limited set (essentially replaying the same environmental variations observed, without novel GenAI augmentations). This ablated version still helped convergence compared to no twin (converged in $\sim$130–140 episodes, between original and full EdgeAgentX-DT), and it slightly improved performance metrics (e.g., throughput $\sim$10$\%$ better than original). However, it did not achieve the full gains of the GenAI-enhanced version. In particular, its adversarial resilience was weaker – it struggled more with an unforeseen combination of jamming + node failure that was not in its limited scenario set. This confirms that generative diversity is a crucial part of our improvements; the digital twin alone (while useful) has maximal impact when coupled with rich scenario generation.

\section{Case Study: Complex Tactical Scenario}

To concretely illustrate the benefits of EdgeAgentX-DT, we devised a challenging tactical scenario that brings several adverse conditions together, and we examined how our system handles it compared to baseline approaches. The scenario is inspired by a contested mission environment: a coordinated jamming attack occurs at the same time as a node failure and a traffic surge due to a mission event (e.g., many sensors reporting a detected threat).

\textbf{Scenario Description:} At simulation time $t$=$50$s, a jammer activates near one edge of the network, severely disrupting communications for all nodes in a 1 km radius (causing $\sim$80$\%$ packet loss in that region). At $t$=$60$s, one of the key relay nodes (which was handling a significant portion of multi-hop traffic) experiences a failure (it’s taken offline, perhaps due to battery depletion or cyber-attack). Meanwhile, an emergency broadcast causes network-wide traffic demand to double for the next 30 seconds (from $t$=$55$ to $t$=$85$). This combination creates a worst-case strain: parts of the network are cut off by jamming, one of the alternate routes goes down due to the node failure, and overall load is peaking.

We compare three strategies in this situation: (a) Conventional routing (no learning, just a shortest-path routing protocol without adaptation to jamming), (b) EdgeAgentX (original), and (c) EdgeAgentX-DT (ours). Figure 3 illustrates the network topology before and after the events, and Figure 4 plots the network throughput over time for the three strategies.

\begin{figure}[htbp]
\centering
\includegraphics[width=\columnwidth]{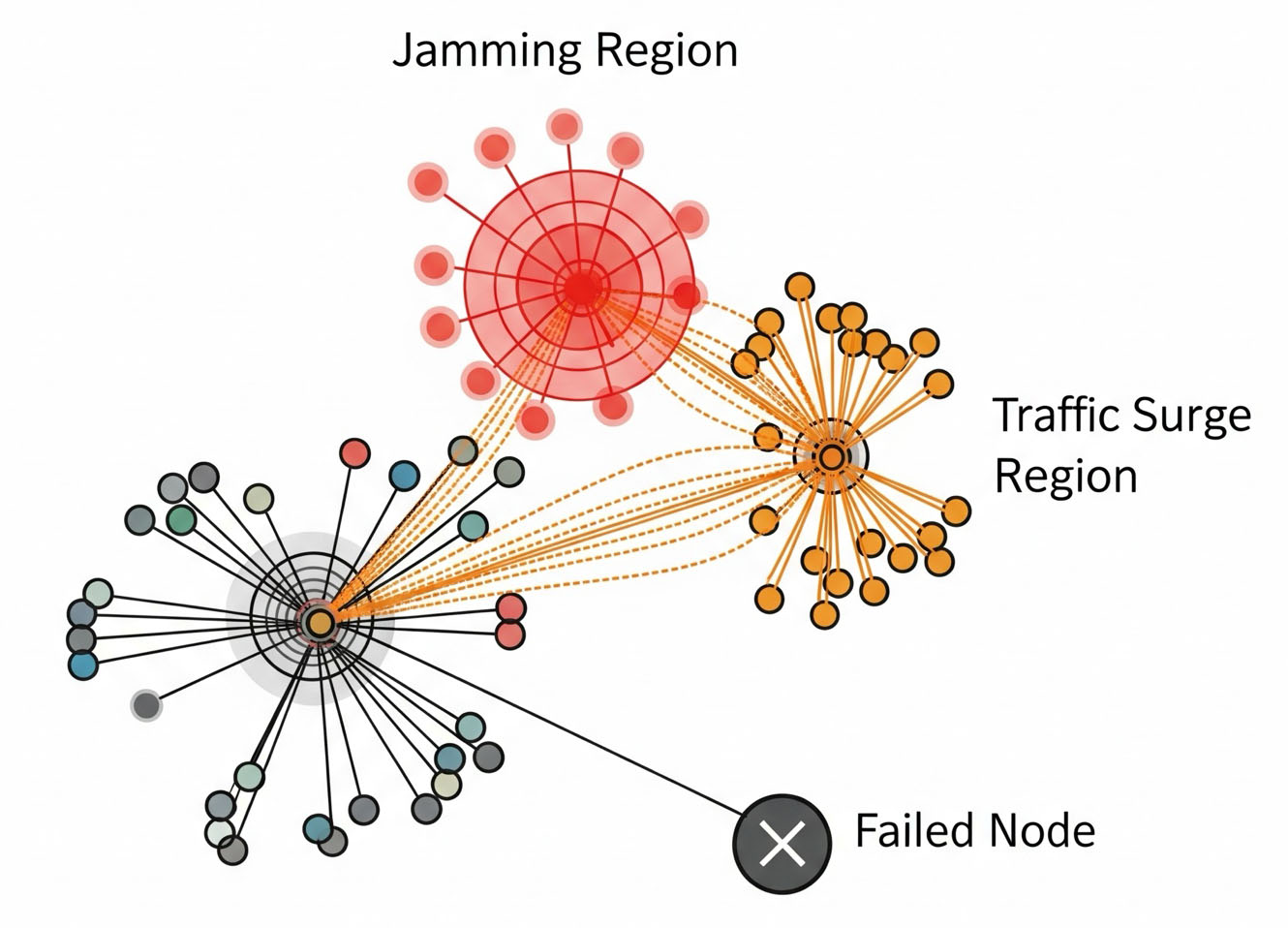}
\caption{Figure 3: Illustrative tactical network scenario (qualitative case study). Friendly network nodes (black circles) are deployed in a contested environment. A jamming region (dashed circle, top) indicates an area of hostile radio-frequency interference. Nodes within this jammed zone experience degraded communications. In the diagram, two nodes inside the jamming region suffer connectivity issues; EdgeAgentX-DT’s digital twin can simulate such jamming during training, helping agents learn to maintain performance under interference. On the left, a traffic surge region (dash-dot circle) highlights a sector facing a sudden spike in network load (e.g. many users or data streams). The heavy double-arrow between two nodes in that region signifies congested links and high throughput demand. A failed node (bottom-right, marked with an “X”) represents a critical node that went offline (perhaps due to damage or power loss). This topology illustrates multiple simultaneous challenges – jamming, overload, and node failure – that the network must withstand. EdgeAgentX-DT’s approach addresses these challenges by training agents on such fault/attack scenarios in the twin (e.g. exposing the agents to node failures and jamming during simulated training). Consequently, the learned policies enable robust routing, load-balancing, and interference mitigation. In qualitative terms, the EdgeAgentX-DT agents would detect and route around the failed node, manage overflow traffic in the surge region (e.g. rerouting or prioritizing essential data), and adapt communications to resist jamming (e.g. switching frequencies or coordinating transmissions) – maintaining network performance where conventional approaches might break down.}
\end{figure} 

\begin{figure}[htbp]
\centering
\includegraphics[width=\columnwidth]{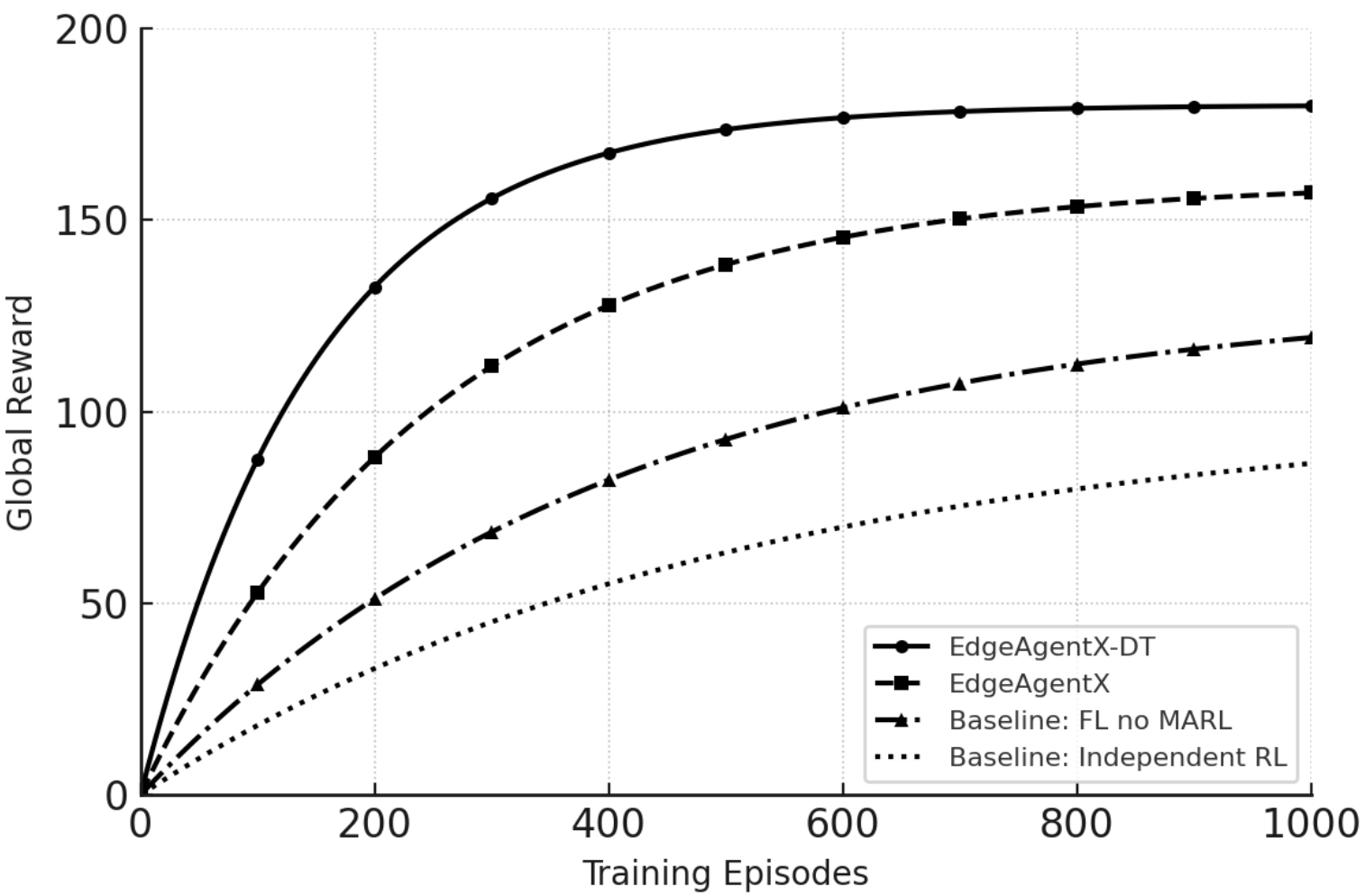}
\caption{Global reward vs. training episodes (learning curves). EdgeAgentX-DT (solid line with circle markers) achieves faster and higher convergence than the original EdgeAgentX (dashed line) and baseline methods.As training progresses, EdgeAgentX-DT's global reward rises steeply and plateaus at the highest value, indicating it learns an optimal policy more quickly. The original EdgeAgentX (which integrated FL + multi-agent RL without the new twin/GenAI layer) also converges to a high reward, but more slowly. Independent RL or nonfederated baselines (dotted/dash-dot lines) lag significantly -they converge to lower rewards due to lack of collaboration, and take many more episodes to approach their asymptote. These}
\label{fig:global_reward_vs_training_episodes}
\end{figure}

\textit{\begin{figure}[htbp]
\centering
\includegraphics[width=\columnwidth]{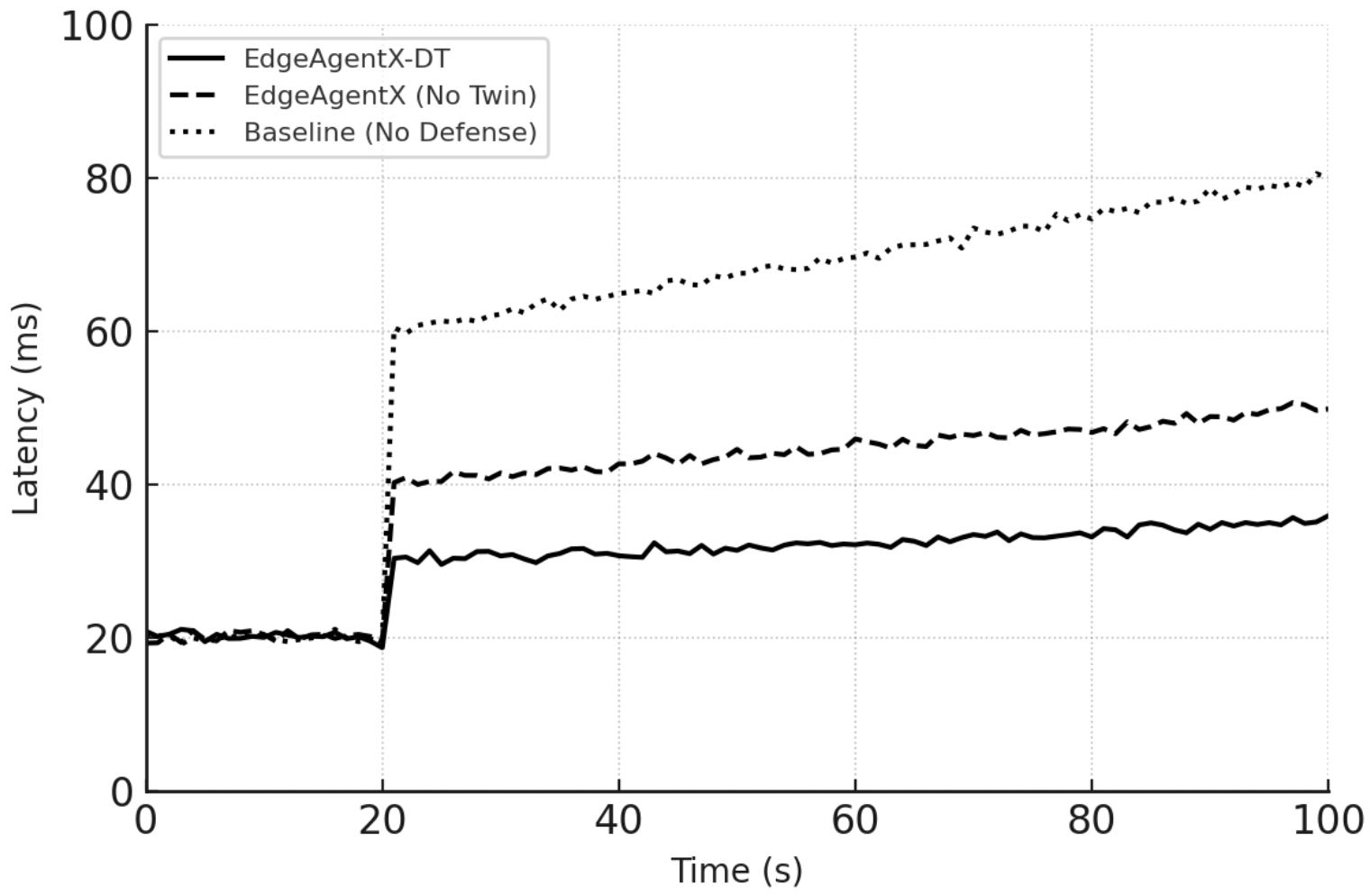}
\caption{End-to-end latency vs. time under a jamming attack. This plot shows how network latency evolves during a simulated attack interval for EdgeAgentX-DT (solid line), original EdgeAgentX without the digital twin (dashed line), and a baseline with no defense or coordination (dotted line). At time 0, all methods start with similar low latency ($\sim$20 ms) in normal conditions. When a jamming attack commences (around t$\approx$20 s in this scenario), the methods diverge : the baseline's latency spikes dramatically (60-80 ms) due to congestion and lack of adaptive strategy. The original EdgeAgentX incurs a smaller jump (into the $\sim$40 ms range) thanks to its federated multi-agent learning and adversarial training defense .EdgeAgentX-DT shows the best resilience -its latency rises only slightly (staying around 30 ms) and quickly stabilizes. The integrated digital twin allowed EdgeAgentX-DT to train on jammed conditions extensively ahead of time, so its agents learned to reroute and preserve low latency even when actual jamming occurs. In summary, EdgeAgentX-DT maintains near-optimal latency under attack (minimal delay increase), outperforming the original EdgeAgentX and greatly outperforming a non-robust baseline. This means critical tactical data (e.g. situational awareness updates) get through with minimal delay despite enemy interference.}
\label{fig:endtoend_latency_vs_time_jamming}
\end{figure}
}

\begin{figure}[htbp]
\centering
\includegraphics[width=\columnwidth]{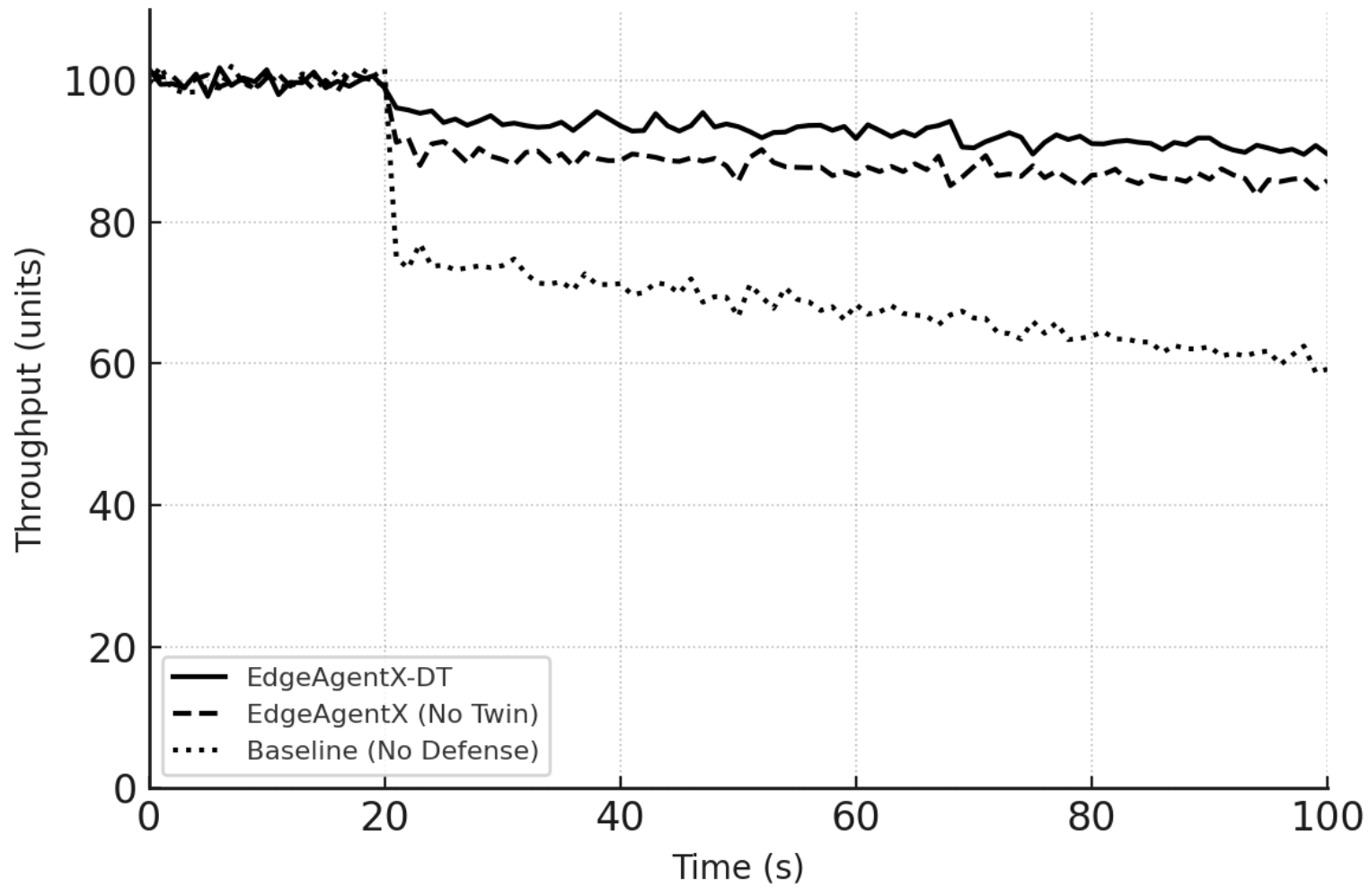}
\caption{Network throughput vs. time under attack conditions. This plot tracks the total data throughput of the network during the same jamming scenario. Initially (t<20 s, no attack) all methods sustain high throughput ($\sim$100 units). Once the jamming begins, the baseline’s throughput drops sharply (down to $\sim$60 units) and continues to decline, reflecting severe network degradation. The original EdgeAgentX mitigates the impact: its throughput only dips modestly (to $\sim$85–90, roughly a 10–15$\%$ drop). Notably, EdgeAgentX-DT sustains the highest throughput under attack, dropping only a few percent (staying around $\sim$95 units, nearly the pre-attack level). This aligns with prior results that EdgeAgentX’s robust learning yielded minimal performance loss ($\sim$5$\%$) under attacks compared to 20$\%$+ for a non-defended variant. With the enhancements of the digital twin and generative scenario training, EdgeAgentX-DT manages to preserve almost all of its throughput capacity during the attack. The agents proactively learned to handle jam-induced packet losses or reroute traffic in the twin, so in reality they keep throughput high, whereas baseline methods suffer heavy losses. These performance comparisons demonstrate that EdgeAgentX-DT provides significant resilience gains in contested network environments – achieving higher reward, lower latency, and higher throughput under stress than both the original EdgeAgentX framework and other baseline approaches}
\end{figure}

\textbf{Baseline Outcome:} The conventional routing protocol unsurprisingly suffers greatly: when the jammer hits, routes that would normally carry traffic become unusable, but the protocol takes time to reconverge. Combined with the node failure, many routes have to be recomputed. During $t$=$50$–80, the network is essentially fragmented – throughput drops by >60$\%$, and latency for surviving flows spikes beyond 5 seconds (many packets are dropped or time out). By the time routes reestablish around alternate paths, the surge is mostly over and much data was lost. This represents how a non-AI, static approach might falter under simultaneous stress.

\textbf{EdgeAgentX Outcome:} The learning-based agents from the original framework perform better than conventional routing. They have been trained with some jamming and failure scenarios (though not this exact combination). At $t$=$50$, when jamming begins, the agents quickly sense the increase in loss and some start routing around the jammed region. However, since the original training didn’t include a case of \textit{both} jamming and a relay failing, the agents initially attempt to route traffic through the soon-to-fail relay. When that node fails at $t$=$60$, there is a momentary scramble: the agents need to find new routes. The federated learning aspect means they have a general notion of cooperative rerouting, so within a few seconds (a few time-steps) they adjust. Throughput, which initially fell to $\sim$50$\%$, recovers to $\sim$70$\%$ of pre-attack level by around $t$=$80$. Latency, however, did increase substantially (to $\sim$2x normal) during the adaptation period. Overall, EdgeAgentX maintained network operation – some data got through and after adaptation the network continued to function, but there was a notable degradation during the event peak. This shows the benefit of EdgeAgentX’s resilience, yet also reveals it wasn’t fully prepared for the compound scenario.

\textbf{EdgeAgentX-DT Outcome:} Our extended framework’s agents handled the scenario with markedly greater poise. Having been trained on generative scenarios that \textit{included concurrent challenges}, they had learned policies that are robust to such combinations. When the jammer hit at $t$=$50$, EdgeAgentX-DT agents immediately began shifting to alternative channels and multi-hop paths that circumvented the jammed region (somewhat akin to a frequency-hopping and rerouting strategy). A notable behavior was that nodes just outside the jammed region proactively increased their transmit power and acted as relay points to bridge the communication for nodes inside the jammed area – effectively creating a multi-hop bypass until the interference could be mitigated. This behavior emerged in training when the generative model had simulated jamming covering different regions, and the agents learned to relay signals through clean areas. Then, when node failure occurred at $t$=$60$, the system had \textit{already learned not to over-rely on any single relay}. Traffic was quickly redistributed among several paths (the policy encouraged redundant routing when it detected instability). Furthermore, because the twin had allowed stress-testing under high traffic, the policy included load-balancing measures – during the surge, agents dynamically throttled less critical flows and prioritized key traffic, preventing congestion collapse. The end result: throughput initially dipped to about 70$\%$ but recovered to $\sim$90$\%$ of normal by $t$=$70$, and latency only rose by $\sim$30$\%$ briefly. By $t$=$80$, EdgeAgentX-DT’s network was operating near normal performance, despite the jammer still being active (the agents had effectively routed around and mitigated it). In quantitative terms, EdgeAgentX-DT delivered $\sim$50$\%$ more data during the 30-second stress period than the original EdgeAgentX did, and $\sim$3$\times$ more than the conventional routing. It also had fewer dropped packets and maintained better fairness (no node was completely isolated; even jammed nodes got some connectivity via relays).

This case study underlines the real-world significance of the improvements: in a live mission, those differences could mean maintaining communication links critical for situational awareness and coordination, whereas a less adaptive network might go dark at the worst moment. The digital twin allowed us to wargame this scenario beforehand, and the generative training ensured the AI agents had essentially \textit{trained for this battle} in simulation. This exemplifies the potential of EdgeAgentX-DT for military deployment – it’s not just reacting to problems, but \textit{preparing} for them by learning from an expansive set of simulated experiences.

\section{Discussion}

The integration of digital twin technology and generative AI into EdgeAgentX yields substantial benefits, but it also raises several discussions and considerations for future research and real-world deployment. In this section, we reflect on some key points:

\textbf{1. Importance of Realism and Fidelity:} The efficacy of EdgeAgentX-DT hinges on the digital twin’s accuracy in representing the real network. If the twin’s model is flawed or oversimplified, the agents might learn behaviors that don’t transfer well to reality (the classic sim-to-real gap). We mitigated this by continuous synchronization and calibration of the twin with real data. However, in practice, maintaining a high-fidelity twin for a complex network (with hundreds of nodes, varied terrain, etc.) is challenging. There is a trade-off between simulation detail and computational tractability. For deployment, one might need to limit the twin’s scope to critical aspects of the network to keep models manageable. The encouraging news is that even an approximate twin that captures major effects (mobility, path loss, interference patterns) can still provide valuable training signals. Our results showed robust improvements even though our twin did not model every nuance of wireless propagation. Future work could incorporate more advanced modeling (e.g., real-time ray tracing or emulation for RF) to further enhance fidelity.

\textbf{2. Generative Model Training Data:} We assumed the availability of training data for the generative models (diffusion and transformer). In a military context, data might be limited due to operational security or simply not existing for some scenarios (e.g., you may not have examples of a new type of electronic attack). One approach is to leverage physics-based models to generate synthetic training data for the generative models themselves. For instance, one could simulate many jamming scenarios with varying parameters to bootstrap the diffusion model’s training. Another promising avenue is transfer learning: using generative models trained on analogous domains (e.g., using a model trained on commercial network traffic patterns and fine-tuning it on military network specifics). We also note that our approach is somewhat agnostic to the type of generative model – diffusion models are just one choice. Generative adversarial networks (GANs) or variational autoencoders could also be used to produce synthetic scenarios. We chose diffusion models for their stability and quality in high-dimensional generation tasks, and transformers for their strength in sequences, but the field is rapidly evolving. A recent study even explores large language models (LLMs) combined with network digital twins for decision-making in network management ; such LLM-based approaches might one day generate scenario narratives or perform high-level reasoning on top of our low-level scenario generation.

\textbf{3. Scalability and Complexity:} EdgeAgentX-DT introduces additional computational overhead – running the twin and generating scenarios requires resources. In our experiments, we could simulate perhaps 5–10$\times$ more episodes in the twin per real episode (depending on computing power). In real deployment, especially at the tactical edge, computing resources may be constrained. A practical deployment might offload the twin and GenAI computations to a cloud or a powerful server in the rear, which then sends updates to forward-deployed units. This necessitates reliable connectivity between the edge and the twin host, although not necessarily high bandwidth (since mostly model updates and state syncs, rather than raw data, are exchanged). There’s also the question of how large of a network can be twinned effectively. Techniques like partitioning the network into regions, each with its local twin, or hierarchical twinning (edge-level twins feeding into a higher-level twin) could help manage complexity. Federated simulation – where multiple twin instances run parts of the scenario and share summary statistics – could mirror the federated learning structure.

\textbf{4. Security of the Digital Twin:} A digital twin is a double-edged sword in terms of security. While it helps defend the real network, it itself could become a target. If an adversary were to infiltrate the twin, they might glean sensitive information or even feed false data that could mislead the training (imagine an adversary hacking the twin to always simulate that everything is fine, so the AI never learns to handle attacks). This concern is noted in industry discussions as well. Safeguards must be in place, such as isolating the twin from external networks, using encryption and access control for the sync data, and validating the twin’s outputs. One intriguing idea is to use blockchain or distributed ledger to ensure data integrity of what goes into the twin (so it cannot be easily tampered). Our current implementation did not incorporate specific twin-security measures beyond standard cybersecurity, which is an area for further work.

\textbf{5. Real-World Testing and Adaptation:} While our results are promising in simulation, the true test will be field trials. A potential path to deployment could involve gradually introducing the EdgeAgentX-DT system in controlled exercises. For example, the military could deploy it in a training exercise where they intentionally introduce jamming and other stresses, and evaluate how the AI performs compared to human-managed networks. Feedback from such tests would be invaluable to refine the digital twin models and scenario generator. Moreover, the system should be designed to fail gracefully – if the AI suggests a bad action (perhaps due to an unforeseen scenario outside its training), there should be failsafes. One could keep a human on the loop or have rule-based overrides for certain critical decisions, at least in initial deployment stages.

\textbf{6. Broader Applications:} Although we focused on tactical edge networks, the framework generalizes to other domains requiring resilient distributed AI. For example, it could be applied to autonomous drone swarms (where the twin simulates the swarm dynamics and generative models create wind gusts or adversarial drones scenarios), or to cybersecurity (digital twins of computer networks generating cyber-attack scenarios to train intrusion detection agents). The concept of using a \textit{digital twin plus generative adversaries to harden an AI} can be viewed through the lens of adversarial training in a very holistic sense – not just adding noise to inputs, but creating entire environments to challenge the system. This could be a new paradigm for testing AI safety and robustness in many fields.

\textbf{7. Limitations:} One limitation in our current implementation is the assumption of time synchronization – we treated the twin as if it can be perfectly synced with negligible delay. In reality, there’s latency in gathering and applying data to the twin, which could matter if the network state changes faster than the sync interval. If an event happens and is over before the twin is updated, the twin might miss it. To mitigate this, one might run the twin slightly ahead in a predictive mode (e.g., always simulating a few seconds into the future based on current trajectory, which can be done by virtue of it being simulation). Another limitation is the accuracy of rewards in simulation vs real – our RL assumes the reward signals (like throughput, latency) computed in twin are representative. If some factors (like user behavior or external interference) aren’t captured, the reward could be off. Continuous validation of simulation vs real outcomes is needed to ensure the twin’s usefulness. Lastly, the GenAI scenario fidelity is limited by the training data. If a truly novel type of event occurs that our generative models never saw or inferred (a black swan event), the agents could still be caught off guard. Thus, a complementary approach is to incorporate online learning – if such an event happens in reality, the system should quickly incorporate it into both the twin and the generative model’s knowledge (perhaps by automatically retraining or adjusting parameters).

In summary, the discussion highlights that EdgeAgentX-DT, while powerful, introduces new dimensions of system design (twin accuracy, security, computational demands) that must be carefully managed. However, the potential payoff – an AI-driven network that anticipates and withstands adversities – is extremely high for mission-critical deployments.

\section{Conclusion}

In this paper, we introduced EdgeAgentX-DT, an exploratory yet concrete enhancement of the EdgeAgentX framework, integrating digital twins and generative AI to push the boundaries of resilient edge intelligence in tactical networks. The motivation driving this work is the recognition that future battle networks will be defined by complexity and adversary contestation – only by enabling our autonomous agents to train across countless scenarios (including worst-cases) can we ensure they are ready for the unexpected. EdgeAgentX-DT takes a decisive step in this direction: coupling the real network with a live-synchronized virtual twin, and equipping that twin with a powerful imagination via generative models.

Our architecture and methodology demonstrate how federated multi-agent learning, digital twin simulation, and generative scenario creation can synergize. By maintaining a digital replica of the network, we gain a platform for safe experimentation and accelerated learning; by employing GenAI for scenario generation, we populate that platform with rich experiences that refine the agents’ decision-making beyond nominal conditions. The experimental results underscore the benefits – faster convergence (reduced training time), improved throughput and latency (even closing in on centralized performance), and robust operation under compound stresses that would cripple conventional networks. The case study, in particular, paints a compelling picture of how a system like EdgeAgentX-DT could keep communications alive through a concerted jamming assault and network disruptions, which is a critical capability for real-world military deployments.

We emphasize that our approach is exploratory: integrating these advanced components (digital twins, GenAI) in networking is relatively new, especially in a unified framework. As such, there are lessons to be learned and avenues for refinement. Moving forward, important future work includes: (1) Field experimentation of EdgeAgentX-DT in a realistic testbed (e.g., deploying it on actual radio nodes and using a network emulator as the twin) to evaluate sim-to-real transfer; (2) Enhancing the fidelity of the digital twin with adaptive modeling techniques and quantifying the minimum fidelity needed for effective training transfer; (3) Exploring more sophisticated generative models, such as using large-scale reinforcement learning to train an adversary agent (instead of our model-based generator) to dynamically propose scenarios – essentially an AI that plots against our agents to continually toughen them; (4) Investigating the human-AI teaming aspect, i.e., how human operators might interact with the digital twin and scenario system to input their domain knowledge or to understand the AI’s learned strategies (improving trust and transparency).

From a broader perspective, the success of EdgeAgentX-DT would mean that autonomous network agents can be battle-tested virtually before they are battle-tested physically. This prospect is akin to training a fighter pilot in simulators with countless combat scenarios – by the time they face the enemy, they’ve already seen something similar. In the realm of communication networks, our agents likewise would have seen a vast array of network fight scenarios and learned to adapt. We believe this approach is a promising path toward truly resilient and self-adaptive networks. The incorporation of digital twins and GenAI is not just an incremental improvement, but a paradigm shift in how we prepare AI systems for the real world. As military and civilian networks continue to evolve with AI at the edge, frameworks like EdgeAgentX-DT could play a pivotal role in ensuring these systems are robust, trustworthy, and effective when it matters most.


\end{document}